\documentclass[letterpaper]{article} 
\usepackage{aaai2026}  
\usepackage{times}  
\usepackage{helvet}  
\usepackage{courier}  
\usepackage[hyphens]{url}  
\usepackage{graphicx} 
\usepackage{natbib}  
\usepackage{caption} 
\usepackage{booktabs}
\usepackage{multirow}
\usepackage{siunitx}
\usepackage{relsize}
\usepackage{amssymb}
\usepackage{pifont}
\usepackage{bm}
\usepackage{subcaption}
\usepackage{amsmath}
\usepackage{cleveref}

\newcommand{\citefmt}[1]{\small(#1)}
\newcommand{\best}[1]{\textbf{#1}}
\newcommand{\secondbest}[1]{\underline{#1}}

\title{
SparseCoop: Cooperative Perception with Kinematic-Grounded Queries
}
\author{
    Jiahao Wang\textsuperscript{\rm 1},
    Zhongwei Jiang\textsuperscript{\rm 2},
    Wenchao Sun\textsuperscript{\rm 1},
    Jiaru Zhong\textsuperscript{\rm 3},
    Haibao Yu\textsuperscript{\rm 4},
    Yuner Zhang\textsuperscript{\rm 5},
    Chenyang Lu\textsuperscript{\rm 1},
    Chuang Zhang\textsuperscript{\rm 1}, 
    Lei He\textsuperscript{\rm 1},
    Shaobing Xu\textsuperscript{\rm 1}
    \thanks{Corresponding authors: Jianqiang Wang and Shaobing Xu (wjqlws@tsinghua.edu.cn, shaobxu@tsinghua.edu.cn)}
    \addtocounter{footnote}{-1},
    Jianqiang Wang\textsuperscript{\rm 1}
    \footnotemark
    \thanks{This work was supported 
    by the National Natural Science Foundation of China for the Science Fund for Creative Research Groups (No. 52221005) and the Key Project (No. 52131201).
    }
}
\affiliations{
    \textsuperscript{\rm 1}Tsinghua University.
    \textsuperscript{\rm 2}Nanyang Technological University.
    \textsuperscript{\rm 3}The Hong Kong Polytechnic University.\\
    \textsuperscript{\rm 4}The University of Hong Kong.
    \textsuperscript{\rm 5}University of Pennsylvania.\\

}

\begin{document}

\maketitle

\begin{abstract}
Cooperative perception is critical for autonomous driving, overcoming the inherent limitations of a single vehicle, such as occlusions and constrained fields-of-view.
However, current approaches sharing dense Bird's-Eye-View (BEV) features are constrained by quadratically-scaling communication costs and the lack of flexibility and interpretability for precise alignment across asynchronous or disparate viewpoints.
While emerging sparse query-based methods offer an alternative, they often suffer from inadequate geometric representations, suboptimal fusion strategies, and training instability.
In this paper, we propose SparseCoop, a fully sparse cooperative perception framework for 3D detection and tracking that completely discards intermediate BEV representations.
Our framework features a trio of innovations:
a kinematic-grounded instance query that uses an explicit state vector with 3D geometry and velocity for precise spatio-temporal alignment;
a coarse-to-fine aggregation module for robust fusion;
and a cooperative instance denoising task to accelerate and stabilize training.
Experiments on V2X-Seq and Griffin datasets show SparseCoop achieves state-of-the-art performance.
Notably, it delivers this with superior computational efficiency, low transmission cost, and strong robustness to communication latency.
Code is available at \url{https://github.com/wang-jh18-SVM/SparseCoop}.
\end{abstract}


\section{Introduction}
\label{sec:introduction}

A robust perception system is fundamental for autonomous driving, but systems confined to a single vehicle are inherently limited by sensor field-of-view constraints, long-range sensing fall-off, and severe occlusions.
These challenges create a critical bottleneck, preventing autonomous systems from achieving the comprehensive awareness necessary for safe deployment in complex scenarios.
To overcome these individual limitations, cooperative perception has emerged as a key paradigm.
By enabling information exchange between multiple agents—such as vehicles (V2V), infrastructure (V2I), and drones (V2D)—it creates a collective sensing system with capabilities far beyond a single vehicle's.
Among various strategies, feature-level fusion is widely researched for its effective balance between preserving rich information and maintaining manageable communication bandwidth~\cite{gaoVehicleRoadCloudCollaborativePerception2024,hanCollaborativePerceptionAutonomous2023}.

\begin{figure}[t]
    \centering
    \includegraphics[width=0.83\columnwidth]{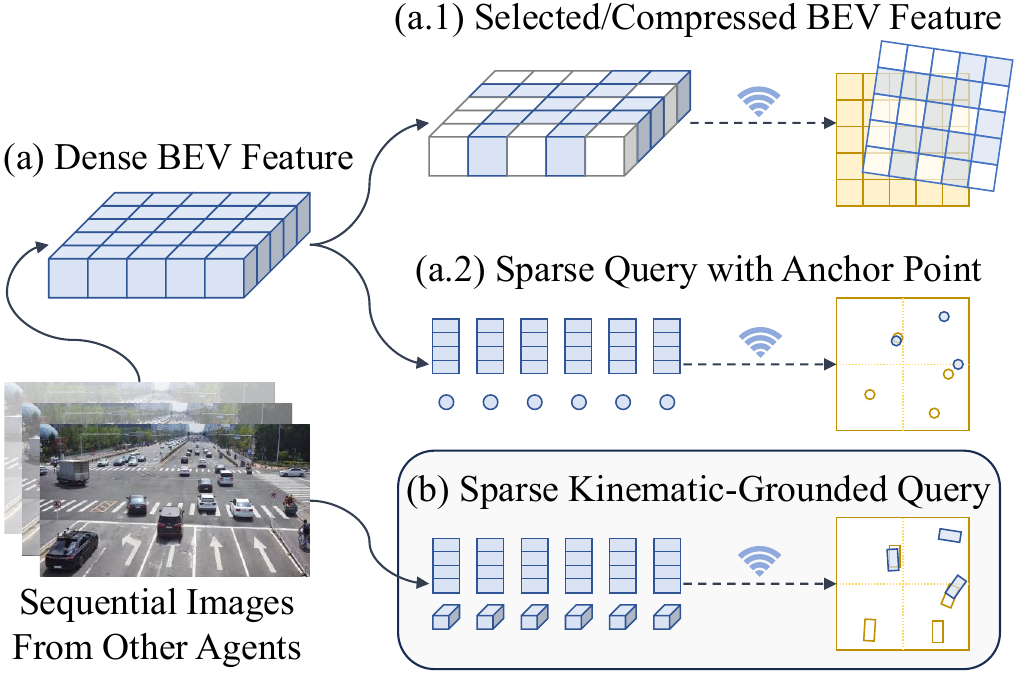}
    \caption{Comparison of cooperative perception pipelines.
    Existing methods (a) are bottlenecked by the intermediate dense BEV features, even when these features are selected, compressed (a.1) or encoded into sparse queries anchored to reference points (a.2).
    In contrast, SparseCoop (b) is a fully sparse paradigm that bypasses BEV step, directly extracting queries grounded by rich state vectors from image features.
    }
    \label{fig:motivation_pipeline}
\end{figure}

The central challenge in feature-level fusion lies in finding a representation that is both efficient for transmission and expressive for robust fusion.
The dominant approach has been to share dense Bird's-Eye-View (BEV) feature maps, which provide a unified spatial grid~\cite{xuOPV2VOpenBenchmark2022,xuCoBEVTCooperativeBirds2023,huCollaborationHelpsCamera2023}, as shown in Figure~\ref{fig:motivation_pipeline}(a.1).
However, this paradigm suffers from fundamental drawbacks: it creates prohibitive communication and computational costs that scale quadratically with perception range, and its abstract scene-level features are difficult to align precisely across agents, especially under temporal asynchrony.
To address these inefficiencies, recent research has shifted towards sparse, query-based methods that represent the scene with a compact set of object-centric queries.

While this emerging sparse paradigm promises greater efficiency and interpretability, it introduces its own set of unresolved challenges.
A primary limitation lies in the geometric representation of the queries themselves.
Often anchored to just a single reference point~\cite{fanQUESTQueryStream2024,yuEndtoEndAutonomousDriving2025}, these queries lack the expressiveness required to handle the significant viewpoint rotations and temporal shifts inherent in real-world cooperative scenarios.
Moreover, their instance fusion strategies are often suboptimal, struggling to effectively integrate information from different agents without losing crucial, unique observations.
The training process is also frequently unstable and inefficient bacause of the limited overlapping viewpoints between agents and sparse supervision signals.
Finally, many of these methods still depend on a dense BEV component~\cite{liuSparseCommEfficientSparse2025,wangIFTRInstanceLevelFusion2025, yuanSparseAlignFullySparse2025, wangCoopDETRUnifiedCooperative2025,zhongCoopTrackExploringEndtoEnd2025}, as shown in Figure~\ref{fig:motivation_pipeline}(a.2), thereby inheriting its computational scaling limitations.

\begin{figure}[t]
    \centering
    \includegraphics[width=\columnwidth]{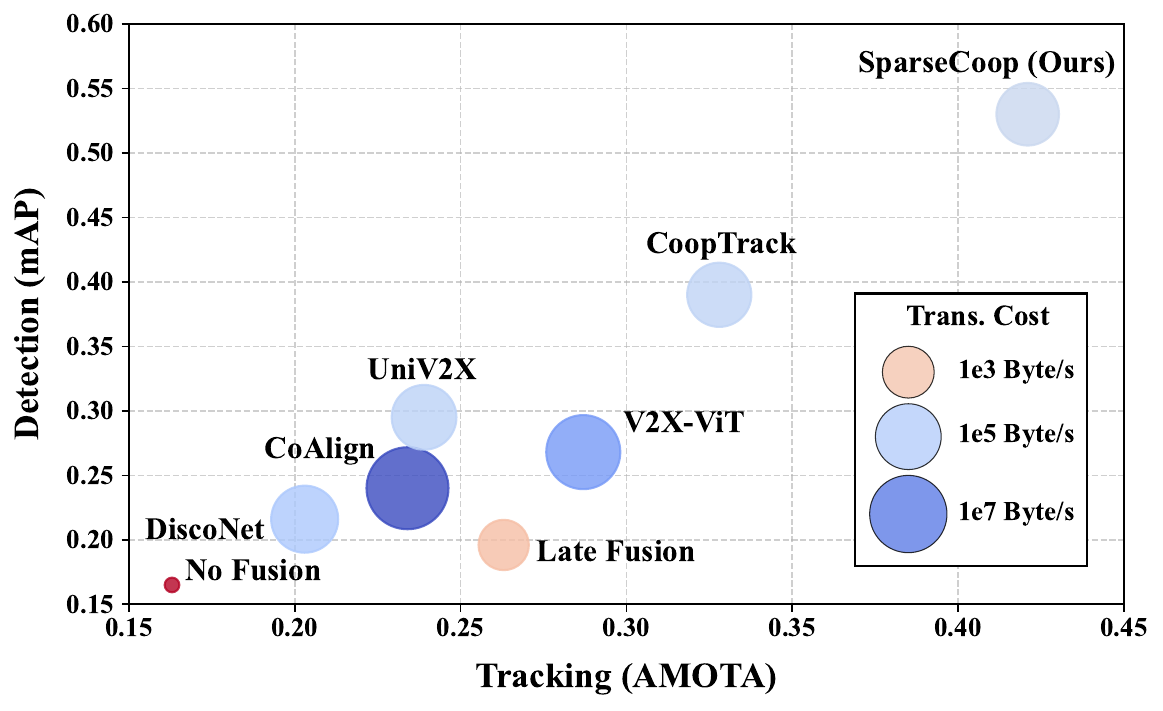}
    \caption{Performance comparison on V2X-Seq dataset.
    The X-axis and Y-axis represent perception metrics, while the bubble size and color encode the transmission cost on a logarithmic scale.
    }
    \label{fig:performance_bubble_chart}
\end{figure}

In this paper, we propose SparseCoop, a fully sparse cooperative detection and tracking framework that directly confronts these challenges.
Our framework is built on a suite of innovations designed for a robust, instance-centric approach:
To address the representation problem, we introduce the kinematic-grounded query, which uses an explicit state vector encoding 3D geometry and velocity, as shown in Figure~\ref{fig:motivation_pipeline}(b), for robust spatio-temporal alignment.
To solve the fusion challenge, we design a coarse-to-fine aggregation module that effectively balances information from both matched and unmatched instances.
Finally, to overcome training instability, we introduce a cooperative instance denoising task that provides a stable and abundant source of supervision.

Our work makes several key contributions:
\begin{itemize}
    \item We propose SparseCoop, a novel, fully sparse cooperative perception framework that operates directly on temporal instance-level representations, eliminating the computational bottlenecks of dense BEV maps.
    \item We introduce a trio of innovations to enable this: a Kinematic-Grounded Association module for precise alignment, a Coarse-to-Fine Aggregation module for effective fusion, and a Cooperative Denoising strategy that stabilizes training.
    \item Our method achieves state-of-the-art (SOTA) detection and tracking performance on both the V2I V2X-Seq and V2D Griffin datasets with low communication and computation costs and strong latency robustness.
\end{itemize}
\section{Related Work}
\label{sec:related}

\begin{figure*}[t]
    \centering
    \includegraphics[width=0.9\textwidth]{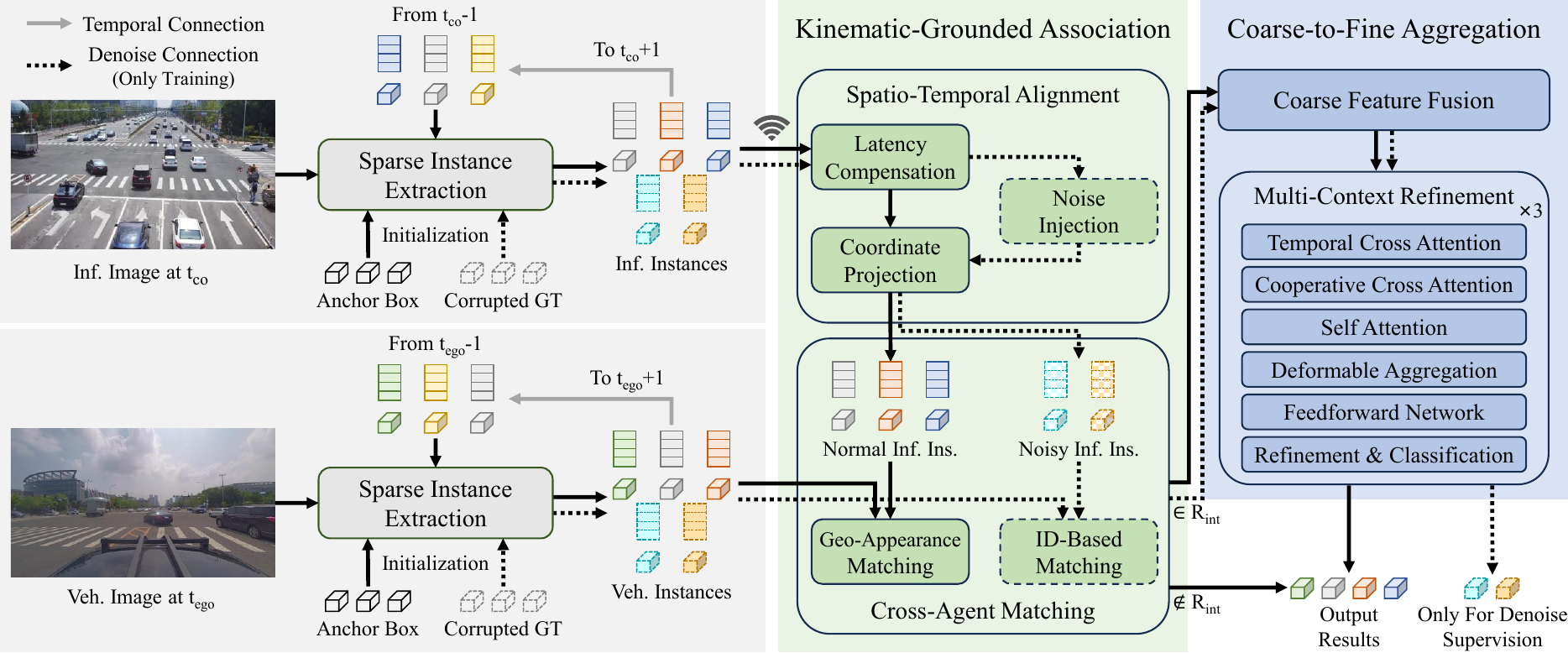}
    \caption{An overview of the SparseCoop framework.
    Each agent independently performs Sparse Instance Extraction.
    The ego-vehicle then uses the proposed Kinematic-Grounded Association and Coarse-to-Fine Aggregation modules to fuse transmitted instances with its own.
    Cooperative Instance Denoising (dashed lines) is only active during training to stabilize convergence.}
    \label{fig:framework}
\end{figure*}



Cooperative perception methods are generally categorized into three fusion strategies~\cite{caillotSurveyCooperativePerception2022, hanCollaborativePerceptionAutonomous2023,gaoVehicleRoadCloudCollaborativePerception2024}: early, intermediate, and late fusion.
Early fusion~\cite{valienteControllingSteeringAngle2019,chenCooperCooperativePerception2019,arnoldCooperativePerception3D2022} directly integrates raw sensor data, offering the potential for the highest performance by retaining rich information, but at the cost of significant bandwidth overhead.
In contrast, late fusion minimizes communication by exchanging final detection results, making cooperation more interpretable but highly dependent on the accuracy of individual perception and coordinate transformation.

Intermediate fusion offers a compromise by transmitting certain network features, with BEV feature maps being the most common representation~\cite{wangV2VNetVehicletovehicleCommunication2020,liLearningDistilledCollaboration2021,huCollaborationHelpsCamera2023, luRobustCollaborative3D2023}.
Due to the high bandwidth required for dense BEV maps, several strategies have been proposed to reduce transmission costs, such as selecting salient regions~\cite{huWhere2commCommunicationefficientCollaborative2022,huCommunicationefficientCollaborativePerception2024,yuanGeneratingEvidentialBEV2023} or applying feature compression techniques~\cite{wangV2VNetVehicletovehicleCommunication2020,xuV2XViTVehicletoeverythingCooperative2022,yinV2VFormerMultiModalVehicletoVehicle2024,wangEMIFFEnhancedMultiscale2024}. However, these scene-level features still lack the flexibility to handle significant spatio-temporal misalignments. For instance, in V2D scenarios with large viewpoint differences, the flattened BEV representation struggles with scale changes and distortion~\cite{wangGriffinAerialGroundCooperative2026}.
What's more, addressing large temporal asynchrony often requires transmitting an additional dense flow map~\cite{yuFlowbasedFeatureFusion2023}, further increasing communication overhead.

To overcome the limitations of dense representations, recent studies have begun to explore the transmission of sparse, instance-level queries~\cite{chenTransIFFInstancelevelFeature2023, fanQUESTQueryStream2024}.
These methods, often built on Transformer architectures~\cite{vaswaniAttentionAllYou2017,wangDETR3D3DObject2022,liuPETRPositionEmbedding2022,liuPETRv2UnifiedFramework2023}, significantly reduce communication costs by only sharing a compact set of object-centric queries.
However, this paradigm faces several challenges.
First, the prevalent query representation—often just a latent feature anchored to a reference point~\cite{fanQUESTQueryStream2024,zhongLeveragingTemporalContexts2024,zhongCoopTrackExploringEndtoEnd2025,liuSparseCommEfficientSparse2025,yuEndtoEndAutonomousDriving2025,wangCoCMTCommunicationEfficientCrossModal2025}—lacks the explicit structure needed for robust spatio-temporal alignment against significant viewpoint and temporal disparities.
Second, existing instance fusion strategies are often suboptimal.
Simpler approaches use linear networks to fuse matched instance pairs~\cite{yuEndtoEndAutonomousDriving2025,zhongCoopTrackExploringEndtoEnd2025}, which is efficient but has limited expressive power.
Other methods employ global~\cite{chenTransIFFInstancelevelFeature2023, fanQUESTQueryStream2024,zhongLeveragingTemporalContexts2024} or selective masked attention~\cite{wangCoCMTCommunicationEfficientCrossModal2025,wangCoopDETRUnifiedCooperative2025} mechanisms across cooperative instances, while powerful, neglecting to fully leverage additional, fine-grained context from the ego-vehicle's own sensor data during the fusion process.
Finally, these methods often struggle with training instability.
Due to differing agent viewpoints and occlusions, the number of co-observed objects available for supervision is inherently low.
This scarcity of positive training samples is compounded by the strict one-to-one matching process required during training, especially in early training stages, which can hinder model convergence.



\section{Method}
\label{sec:method}

\subsection{Preliminaries: Task and Query Definition}
\label{sec:preliminaries}

We adopt the standard task formulation from prominent cooperative perception benchmarks~\cite{yuV2XseqLargescaleSequential2023,wangGriffinAerialGroundCooperative2026}. 
The objective is to generate a temporally consistent set of 3D tracked objects $\{\hat{o}_i\}$. An ego-agent at its current timestamp $t_{\text{ego}}$ utilizes its own sensor data alongside information shared by a cooperative agent from a potentially asynchronous timestamp $t_{\text{co}}$, where $t_{\text{co}} \le t_{\text{ego}}$ due to communication latency. The final output is generated within the ego-agent's predefined region of interest (ROI), $R_{\text{ego}}$.

The fundamental unit of representation, transmission, and fusion in our framework is the \textbf{Kinematic-Grounded Query (KGQ)}. A KGQ is an \textit{instance} defined as a pair $\{\mathcal{F}, \mathcal{S}\}$, where $\mathcal{F}$ is a latent feature vector encoding semantic information and $\mathcal{S}$ is its explicit 11-dimensional state vector, defined as:
\begin{equation}
    \mathcal{S} = (x, y, z, l, w, h, \sin(\theta), \cos(\theta), v_x, v_y, v_z)
\end{equation}
This state vector describes the object's 3D position, dimensions, heading angle, and velocity.
This rich, explicit representation is a key distinction from prior works that rely on simpler geometric inputs like a single reference point, and it is central to our robust alignment and fusion strategy.

With a road side unit (RSU) as an example of a cooperative agent, the overall data flow of our framework is illustrated in Figure~\ref{fig:framework}.
Each agent first independently generates a set of KGQs from its own sensor data. High-confidence instances from the cooperative agent are then transmitted to the ego-vehicle. Upon reception, the ego-vehicle employs the Kinematic-Grounded Association (KGA) and Coarse-to-Fine Aggregation (CFA) modules to process these incoming KGQs, fusing them with its own to produce the final set of tracked objects. During training, additional KGQs are initialized from corrupted ground truth (GT) boxes for Cooperative Instance Denoising (CID).

\subsection{Sparse Instance Extraction}
\label{subsec:sparse-instance-extraction}

As depicted in Figure~\ref{fig:framework}, the first stage of our pipeline, deployed independently on each agent, is Sparse Instance Extraction.
For this, we adapt the Sparse4D framework~\cite{linSparse4DMultiview3D2023,linSparse4DV2Recurrent2023,linSparse4DV3Advancing2023}. 
The process begins by initializing a set of KGQs with predefined anchor boxes derived from dataset priors.
These KGQs are then refined by directly aggregating information from multi-scale image features using deformable attention.
This fully-sparse paradigm is a core design choice, as it bypasses computationally expensive dense BEV representations and their prohibitive scaling costs, making it well-suited for long-range cooperative perception.
To ensure temporal consistency for tracking, the framework employs a recurrent mechanism where high-confidence KGQs are assigned a tracking ID and propagated to the subsequent frame.
This stage ultimately provides a continuous stream of refined, temporally-aware KGQs for the cooperative fusion modules.

\subsection{Kinematic-Grounded Association}
\label{subsec:kinematic-grounded-association}

Once the ego-vehicle receives KGQs from a cooperative agent, it performs association.
This critical step addresses the first challenge outlined in our introduction: robustly matching instances across different viewpoints and asynchronous timestamps.
Existing methods~\cite{yuEndtoEndAutonomousDriving2025,zhongLeveragingTemporalContexts2024} often rely on the Euclidean distance between simple reference points, a strategy that is fragile in complex scenarios.
Our association mechanism overcomes this by leveraging the rich geometric and kinematic information encoded in each KGQ's state vector $\mathcal{S}$, enabling precise alignment and matching.

\noindent{\textbf{Spatio-Temporal Alignment.}}
The first step of association is aligning the transmitted cooperative instances, $\{\mathcal{F}_{\text{co}}(t_{\text{co}}), \mathcal{S}_{\text{co}}(t_{\text{co}})\}$, to the ego-vehicle's current spatio-temporal frame.
We address latency compensation and coordinate projection separately.

\begin{figure}[bt]
    \centering
    \includegraphics[width=0.5\columnwidth]{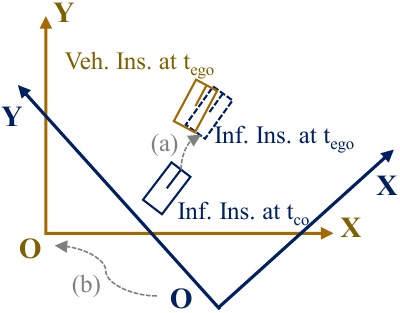}
    \caption{Spatio-Temporal Alignment for KGQ state vectors}
    \label{fig:spatio-temporal-alignment}
\end{figure}

To handle latency from asynchronous communication, we utilize the velocity $(v_x, v_y, v_z)$ encoded within the state vector $\mathcal{S}_{\text{co}}(t_{\text{co}})$.
Applying a constant velocity motion model, we predict the object's state at the ego-vehicle's timestamp, yielding an updated state vector $\mathcal{S}_{\text{co}}(t_{\text{ego}})$, as shown in Figure~\ref{fig:spatio-temporal-alignment}(a).
For the feature vector $\mathcal{F}_{\text{co}}$, we find that the recurrent nature of the extraction module provides sufficient time invariance, so we use it directly: $\mathcal{F}_{\text{co}}(t_{\text{ego}}) = \mathcal{F}_{\text{co}}(t_{\text{co}})$.

Next, we project each instance into the ego-vehicle's coordinate system. The transformation matrix is calculated as $ \mathbf{T}_{\text{co} \to \text{ego}}(t_{\text{ego}}) = \mathbf{T}_{\text{ego} \to \text{glb}}(t_{\text{ego}})^{-1} \cdot \mathbf{T}_{\text{co} \to \text{glb}}(t_{\text{co}}) $.
We apply this to the state vector to get the fully aligned $\widetilde{\mathcal{S}_{\text{co}}}(t_{\text{ego}})$, as shown in Figure~\ref{fig:spatio-temporal-alignment}(b).
For the feature vector, we follow prior work~\cite{fanQUESTQueryStream2024,yuanSparseAlignFullySparse2025} and use a rotation-aware multi-layer perceptron (MLP) to update it:
$$ 
\widetilde{\mathcal{F}_{\text{co}}}(t_{\text{ego}}) = \text{MLP}([\mathcal{F}_{\text{co}}(t_{\text{ego}}); \mathbf{r}_{\text{co} \to \text{ego}}(t_{\text{ego}})])
$$
where $\mathbf{r}_{\text{co} \to \text{ego}}(t_{\text{ego}}) \in \mathbb{R}^{1 \times 9}$ is the flattened rotation matrix from $\mathbf{T}_{\text{co} \to \text{ego}}(t_{\text{ego}})$.

\noindent{\textbf{Cross Agent Matching}}
After alignment, we associate the ego-vehicle's instances $\{\mathcal{F}_{\text{ego}}, \mathcal{S}_{\text{ego}}\}$ with the aligned cooperative instances $\{\widetilde{\mathcal{F}_{\text{co}}}, \widetilde{\mathcal{S}_{\text{co}}}\}$.
For simplicity, we omit the timestamp $t_{\text{ego}}$ in the following sections.

First, cooperative instances outside the ego-vehicle's ROI $R_{\text{ego}}$ are filtered out, as they are not involved in the task.
Within $R_{\text{ego}}$, we further define a smaller interaction range, $R_{\text{int}}$, to determine which instances undergo fusion.
This distinction is based on the principle that for distant or occluded objects, the ego-vehicle's own sensory data is often unreliable.
Forcing fusion in such cases can corrupt high-quality data from cooperative agents with noisy local estimates.
Therefore, we only perform matching and subsequent fusion for instances within this trusted near-field region, $R_{\text{int}}$.

For instances inside $R_{\text{int}}$, we introduce a \textit{Geo-Appearance Matching} (GAM) strategy to find optimal instance pairs.
This approach directly addresses the fragility of prior methods~\cite{zhongLeveragingTemporalContexts2024,yuEndtoEndAutonomousDriving2025} that rely on a simple Euclidean distance between single reference points, which can be ambiguous in dense traffic scenarios.
To improve robustness, our GAM strategy constructs a pairwise cost matrix $C$ using two distinct components.
For an ego-instance $i$ and a cooperative-instance $j$, a \textit{Geometric Similarity} is computed as a weighted L1 distance between their state vectors ($\mathcal{S}_{\text{ego},i}$ and $\widetilde{\mathcal{S}_{\text{co},j}}$), and an \textit{Appearance Similarity} is computed as the cosine distance between their feature vectors ($\mathcal{F}_{\text{ego},i}$ and $\widetilde{\mathcal{F}_{\text{co},j}}$).
The final cost combines these two scores, creating a more discriminative matching criterion that ensures reliable associations, especially for closely-located objects.

The association process results in three distinct groups of instances: 1) the successfully matched pairs, 2) the unmatched ego-vehicle instances, and 3) the unmatched cooperative instances.
This entire collection is then passed to the next module for comprehensive refinement.

\subsection{Coarse-to-Fine Aggregation}
\label{subsec:coarse-to-fine-aggregation}

This module 
consists of a coarse fusion step followed by a more intensive refinement process, as detailed below.

\paragraph{Coarse Fusion.}
For the successfully matched instance pairs, we first perform a coarse fusion. Following prior works~\cite{yuEndtoEndAutonomousDriving2025,zhongCoopTrackExploringEndtoEnd2025}, we employ a lightweight linear network to fuse their respective feature vectors, creating a single, consolidated feature representation for each matched object.
$$
\mathcal{F}_{\text{fused}} = \text{MLP}([\mathcal{F}_{\text{ego}}; \widetilde{\mathcal{F}_{\text{co}}}])
$$

\paragraph{Multi-Context Refinement.}
The resulting fused KGQs, along with all unmatched ones from both the ego-vehicle and the cooperative agent, then proceed through an iterative refinement process.
This process is inspired by the Sparse4D decoder architecture~\cite{linSparse4DMultiview3D2023,linSparse4DV2Recurrent2023,linSparse4DV3Advancing2023} but is specifically adapted for the cooperative perception task.
Unlike prior cooperative methods that often refine instances using only temporal~\cite{zhongLeveragingTemporalContexts2024} or cooperative contexts~\cite{wangCoCMTCommunicationEfficientCrossModal2025,wangCoopDETRUnifiedCooperative2025}, we argue that leveraging the ego-vehicle's rich image features is also critical for cooperation.

Each refinement stage is composed of several key operations to leverage multiple contexts.
First, \textit{Multi-Head Attention} mechanisms enable rich instance-level interactions.
Temporal cross-attention links current instances with those from the previous frame, enabling the model to understand object motion and maintain tracking consistency.
Cooperative cross-attention allows interaction with the full set of aligned cooperative KGQs, which is essential for incorporating information from areas occluded to the ego-vehicle.
Self-attention captures relationships among all instances within the current frame, helping the model reason about the scene's layout and avoid duplicate detections.
For these operations, the state vectors $\mathcal{S}$ are transformed into high-dimensional embeddings to serve as positional encodings.

Following the attention layers, a \textit{Deformable Aggregation} module further refines each instance by sampling from the ego-vehicle's multi-scale image features.
This step is crucial for grounding the abstract instance representations in the raw visual data, leading to more precise localization.
Finally, a \textit{Prediction Head}, consisting of a feedforward network and an output layer, predicts the classification scores and state vector refinements.

\subsection{Cooperative Instance Denoising}
\label{subsec:cooperative-denoising}

\begin{figure}[t]
    \centering
    \begin{subfigure}[b]{0.5\columnwidth}
        \centering
        \includegraphics[width=\linewidth]{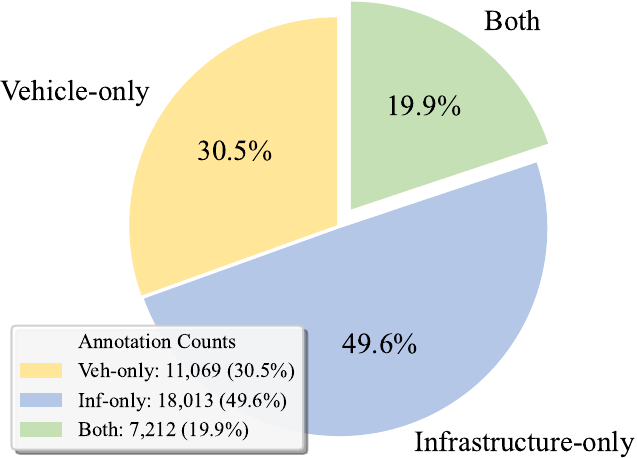} 
        \caption{Distribution of annotation visibility in V2X-Seq dataset}
        \label{fig:annotation_visibility}
    \end{subfigure}
    \begin{subfigure}[b]{0.47\columnwidth}
        \centering
        \includegraphics[width=\linewidth]{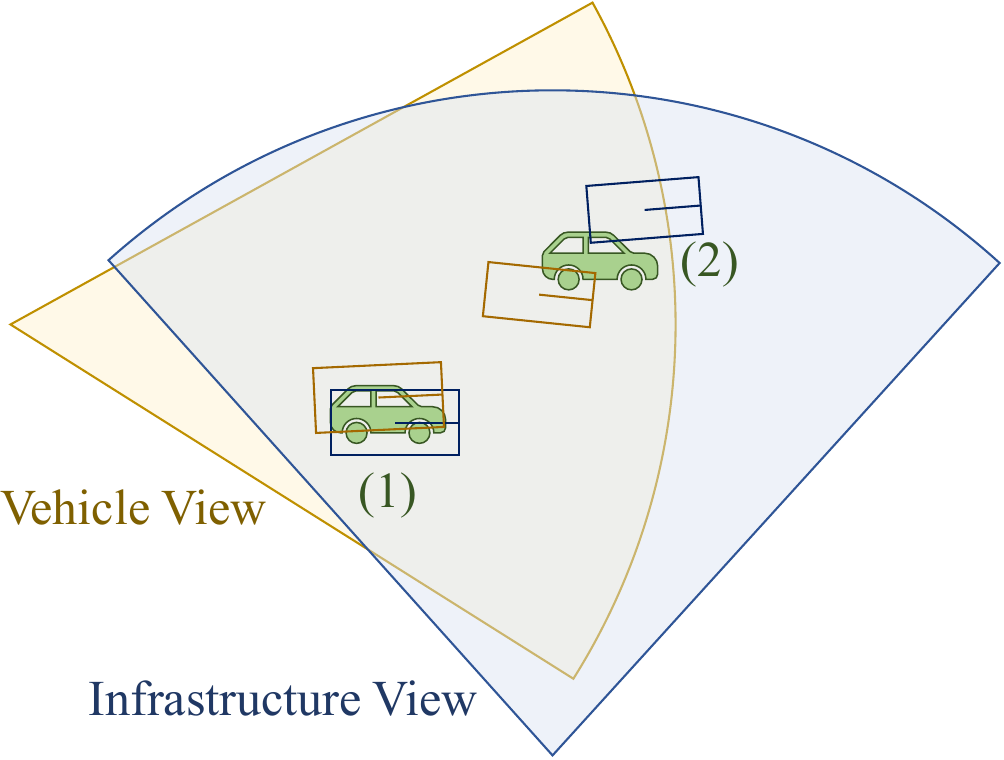} 
        \caption{KGQs from different agents for the same GTs (as car icons)}
        \label{fig:kgq_matching}
    \end{subfigure}
    \caption{Motivation for CID. (a) A significant portion of ground-truth objects are visible to only one agent, limiting opportunities for cooperative supervision. (b) Even when an object is visible to both agents, predictions for the same GT (2) can be too far apart to be matched, further reducing positive samples for the fusion module.}
    \label{fig:cid_motivation}
\end{figure}

A key challenge in sparse cooperative perception is the scarcity of positive supervision signals, particularly during the early stages of training when few instances are successfully matched across agents, as illustrated in Figure~\ref{fig:cid_motivation}. 
To address this, we draw inspiration from denoising techniques in object detection~\cite{linSparse4DV3Advancing2023,liMaskDINOUnified2023,liDNDETRAccelerateDETR2024} and introduce a Cooperative Instance Denoising task.
The denoising instances are initialized by adding small perturbations to GT objects and then fed into the network alongside the normal ones.
Since the denoising queries for each agent originate from the same GT set, their correspondence is known a priori.
This design provides a stable and abundant supervisory signal that guides the fusion module, promoting robust model convergence from the outset of training.

\noindent{\textbf{Noise Injection.}}
To simulate realistic uncertainties, we inject two types of noise into the GT state vectors during training.
The first, \textit{Observation Noise}, simulates intra-agent uncertainties like sensor measurement errors by adding small, random perturbations to the attributes of GT state vectors in their respective local coordinate systems.
Specifically, we add noise sampled from a uniform distribution over (-2.0m, 2.0m) for positional attributes and (-0.5, 0.5) for all other dimensions.
The second, a novel \textit{Transformation Noise}, simulates inter-agent uncertainties such as extrinsic calibration errors or timestamp asynchrony.
This is achieved by applying minor random rotations and translations to the coordinate transformation matrix $\mathbf{T}_{\text{co} \to \text{ego}}$.
The translations are sampled from a normal distribution with a mean of 0 and a standard deviation of 1.0m, while rotations are sampled with a standard deviation of 2.0 degrees.

\noindent{\textbf{Denoising Pipeline and Supervision.}}
The denoising instances are processed through a pipeline that mirrors the main network but incorporates critical modifications to ensure effective and clean supervision.
First, during cross-agent matching, denoising queries are associated using their tracking IDs inherited from the original GT objects.
This provides the network with a stable and plentiful stream of perfectly matched pairs, which is essential for learning the feature alignment and fusion process.
Second, to prevent information leakage, we enforce a strict separation between the normal and denoising pipelines within the attention mechanisms of the refinement module.
A custom attention mask is implemented to ensure normal KGQs and denoising queries operate in distinct groups.
This separation is vital as it prevents the model from accessing GT information during the standard perception task, forcing it to learn meaningful fusion strategies instead of trivial shortcuts.
The entire process provides strong, stable supervisory signals throughout all training stages, leading to more robust convergence.


\begin{table*}[t]
    \centering

    \begin{tabular}{c ccc cccc}
        \toprule
        \multirow{2}{*}[-0.5ex]{\textbf{Method}} & \multicolumn{3}{c}{\textbf{V2X-Seq}} & \multicolumn{4}{c}{\textbf{Griffin-25m}} \\
        \cmidrule(lr){2-4} \cmidrule(lr){5-8}
        & \textbf{AP} $\uparrow$ & \textbf{AMOTA} $\uparrow$ & \textbf{TC (BPS)} $\downarrow$ & \textbf{AP} $\uparrow$ & \textbf{AMOTA} $\uparrow$ & \textbf{TC (BPS)} $\downarrow$ & \textbf{CE (FPS)} $\uparrow$ \\
        \midrule

        No Fusion Baseline & 0.166 & 0.130 & 0 & 0.375 & 0.365 & 0 & 8.10 \\
        Late Fusion Baseline & 0.196 & 0.263 & $6.60 \times 10^2$ & 0.378 & 0.377 & $1.56 \times 10^3$ & 6.83 \\
        Early Fusion Baseline & 0.243 & 0.209 & $\num{8.19e7}$ & 0.607 & 0.670 & $\num{3.11e8}$ & 5.17 \\
        \midrule

        V2X-ViT \citefmt{ECCV 2022} & 0.268 & 0.287 & $\num{2.56e6}$ & 0.465 & \secondbest{0.508} & $\num{8.00e5}$ & 7.56 \\
        Where2Comm \citefmt{NIPS 2022} & 0.162 & 0.106 & $\num{5.40e5}$ & 0.396 & 0.406 & $\num{3.30e5}$ & \secondbest{7.60} \\
        UniV2X \citefmt{AAAI 2025} & 0.295 & 0.239 & $\num{6.96e4}$ & 0.419 & 0.456 & \bm{$5.58 \times 10^4$} & 7.06 \\
        CoopTrack \citefmt{ICCV 2025} & \secondbest{0.390} & \secondbest{0.328} & \secondbest{$\num{5.64e4}$} & \secondbest{0.479} & 0.488 & $\num{1.17e5}$ & 6.23 \\
        SparseCoop \citefmt{Ours} & \best{0.530} & \best{0.421} & \bm{$3.17 \times 10^4$} & \best{0.559} & \best{0.509} & \secondbest{$\num{9.73e4}$} & \best{11.64} \\
        \bottomrule
    \end{tabular}
    \caption{
        Model performance, transmission cost (TC), and computational efficiency (CE) comparison on the V2X-Seq and Griffin-25m datasets.
        Baseline performance metrics are sourced from~\cite{zhongCoopTrackExploringEndtoEnd2025,wangGriffinAerialGroundCooperative2026}.
        Bold and underlined values denote the best and second-best performance among intermediate fusion methods.
    }
    \label{tab:performance}
\end{table*}

\section{Experiments}
\label{sec:experiments}

We evaluate our approach on two benchmark datasets: V2X-Seq~\cite{yuV2XseqLargescaleSequential2023} and Griffin~\cite{wangGriffinAerialGroundCooperative2026}.
We present the results of quantitative experiments in this section, while implementation details and qualitative visualizations are included in the appendix.

\subsection{Datasets and Metrics}
\label{subsec:datasets-and-metrics}

\noindent\textbf{V2X-Seq.}
This is a large-scale, sequential dataset designed for V2I cooperative 3D object detection and tracking~\cite{yuV2XseqLargescaleSequential2023}.
It features sensor data from both an ego-vehicle and a connected RSU, capturing complex urban traffic scenarios with significant occlusions.
We follow the official protocol in its CVPR 2025 challenge~\cite{haoResearchChallengesProgress2025} and report performance on the 2Hz validation split.

\noindent\textbf{Griffin.}
This is a pioneering simulated dataset for aerial-ground cooperative perception~\cite{wangGriffinAerialGroundCooperative2026}.
It presents unique challenges due to the large viewpoint disparity and dynamic transformations in V2D scenarios, and provides a code
interface to inject real-world imperfections during training and inference, including localization errors, communication latency, and packet loss, enabling a full evaluation of model robustness.
We report results on the validation split of \textit{Griffin-25m} subset.

\begin{table*}[!t]
    \centering
    \newcolumntype{C}{>{\centering\arraybackslash}m{1.8cm}}
    
    \begin{tabular}{*{6}{C} S[table-format=1.3] S[table-format=1.3]}
    \toprule
    \multicolumn{2}{c}{\small{Kinematic-Grounded Association}} & \multicolumn{2}{c}{\small{Coarse-to-Fine Aggregation}} & \multicolumn{2}{c}{\small{Cooperative Denoising}} & \multicolumn{2}{c}{\small{Metrics}} \\
    \cmidrule(lr){1-2} \cmidrule(lr){3-4} \cmidrule(lr){5-6} \cmidrule(lr){7-8}
    LC & GAM & CFF & MCR & ON & TN & {AP$\uparrow$} & {AMOTA$\uparrow$} \\
    \midrule
    \checkmark & \checkmark & \checkmark & \checkmark & \checkmark & \checkmark & \secondbest{0.530} & \textbf{0.421} \\
    \midrule
    \ding{55}  & \checkmark & \checkmark & \checkmark & \checkmark & \checkmark & 0.505 & 0.414 \\
    \checkmark & \ding{55}  & \checkmark & \checkmark & \checkmark & \checkmark & 0.502 & 0.414 \\
    \ding{55}  & \ding{55}  & \checkmark & \checkmark & \checkmark & \checkmark & 0.505 & 0.408 \\
    \midrule
    \checkmark & \checkmark & \ding{55}  & \checkmark & \checkmark & \checkmark & 0.489 & 0.375 \\
    \checkmark & \checkmark & \checkmark & \ding{55}  & \checkmark & \checkmark & 0.512 & 0.379 \\
    \midrule
    \checkmark & \checkmark & \checkmark & \checkmark & \ding{55}  & \checkmark & 0.521 & \secondbest{0.416} \\
    \checkmark & \checkmark & \checkmark & \checkmark & \checkmark & \ding{55}  & \best{0.531} & 0.394 \\
    \checkmark & \checkmark & \checkmark & \checkmark & \ding{55}  & \ding{55}  & 0.521 & 0.352 \\
    \bottomrule
    \end{tabular}
    \caption{Ablation study of our proposed modules on the V2X-Seq dataset.
    The top row shows the full model.
    Subsequent rows show performance when components are removed (\ding{55}).
    Experiments are grouped by the major submodule being ablated.
    Abbreviations are: LC (Latency Compensation), GAM (Geo-Appearance Matching), CFF (Coarse Feature Fusion), MCR (Multi-Context Refinement), ON (Observation Noise), and TN (Transformation Noise).
    }
    \label{tab:ablation}
\end{table*}

\noindent\textbf{Evaluation Metrics.}
Both benchmarks adopt established metrics from the NuScenes benchmark~\cite{caesarNuScenesMultimodalDataset2020} for 3D object detection and tracking, including Average Precision (AP) to assess detection quality and Average Multi-Object Tracking Accuracy (AMOTA) for tracking performance.
We also assess the transmission cost in Bytes per second (BPS) and computational efficiency by inference Frames per second (FPS).

\subsection{Comparison with Existing Works}

The detection and tracking performance of SparseCoop across the two datasets are presented in Table~\ref{tab:performance}, which details the AP and AMOTA scores alongside the associated communication costs.
Our method achieves a new SOTA detection and tracking performance on both datasets, with a competitive transmission cost.
On the V2X-Seq dataset, SparseCoop achieves an AP of 0.530 and an AMOTA of 0.421, outperforming all other methods.
Notably, this performance is achieved with the lowest transmission cost among all learning-based methods, at only $3.17 \times 10^4$ BPS.
This represents a significant improvement over the next best-performing methods, CoopTrack and UniV2X, while requiring substantially less bandwidth.
On the Griffin-25m dataset, SparseCoop continues to demonstrate its superiority, achieving the highest AP (0.559), AMOTA (0.509) and computational efficiency (11.64 FPS).
While its transmission cost of $9.73 \times 10^4$ BPS is second only to UniV2X, the performance gains are substantial, with 8\% in AP and over 50\% in FPS against the next-best competitor.



\subsection{Ablation Study}

To validate the effectiveness of our proposed components, we conduct a comprehensive ablation study on the V2X-Seq dataset, systematically dissecting the contributions of the Kinematic-Grounded Association, Coarse-to-Fine Aggregation, and Cooperative Instance Denoising modules. The detailed results are presented in Table~\ref{tab:ablation}.

\noindent\textbf{Effect of Kinematic-Grounded Association.}
The exclusion of Latency Compensation (LC), which rectifies temporal asynchrony using kinematic priors, results in a performance drop to 0.505 AP and 0.414 AMOTA.
A similar degradation is observed when we remove KGQ's effect on Geo-Appearance Matching (GAM) and revert to a simpler point-based matching metric, with scores falling to 0.502 AP and 0.414 AMOTA.
This underscores that leveraging the full geometric and kinematic information encoded in KGQ state vectors is critical for accurate cross-agent instance association.
When both components are disabled, the model's performance deteriorates further, confirming their synergistic contribution to achieving robust alignment in complex cooperative scenarios.

\noindent\textbf{Effect of Coarse-to-Fine Aggregation.}
The Coarse-to-Fine Aggregation module is designed to effectively fuse information from matched instances and refine their representations.
Removing the initial Coarse Feature Fusion (CFF) stage, where matched pairs are first integrated, causes a substantial decline in performance to 0.489 AP and 0.375 AMOTA.
This highlights the necessity of this direct fusion step for consolidating redundant observations.
Furthermore, ablating the subsequent Multi-Context Refinement (MCR) process leads to a significant performance drop to 0.512 AP and 0.379 AMOTA.
This comprehensive stage facilitates interaction with temporal and cooperative instances while also using deformable aggregation to ground queries in the ego-vehicle's image features.
The performance decline underscores that both multi-agent contextualization and refinement against raw visual data are crucial for comprehensive scene understanding.

\noindent\textbf{Effect of Cooperative Denoising.}
This training-only task is introduced to provide stable and abundant supervisory signals, mitigating the challenges of sparse positive samples in cooperative settings.
Removing Observation Noise (ON), which simulates intra-agent sensor uncertainty, slightly degrades performance.
However, the removal of Transformation Noise (TN), which models inter-agent calibration and asynchrony errors, leads to a more significant drop in tracking performance (AMOTA from 0.421 to 0.394), confirming its importance for learning robust alignment.
When the entire denoising pipeline is deactivated, the model suffers a severe degradation, particularly in tracking, with AMOTA plummeting to 0.352.
This result decisively demonstrates that the cooperative denoising task is crucial for stabilizing the training process and enabling the model to learn effective fusion strategies.

\begin{figure}[t]
    \centering
    \includegraphics[width=0.7\columnwidth]{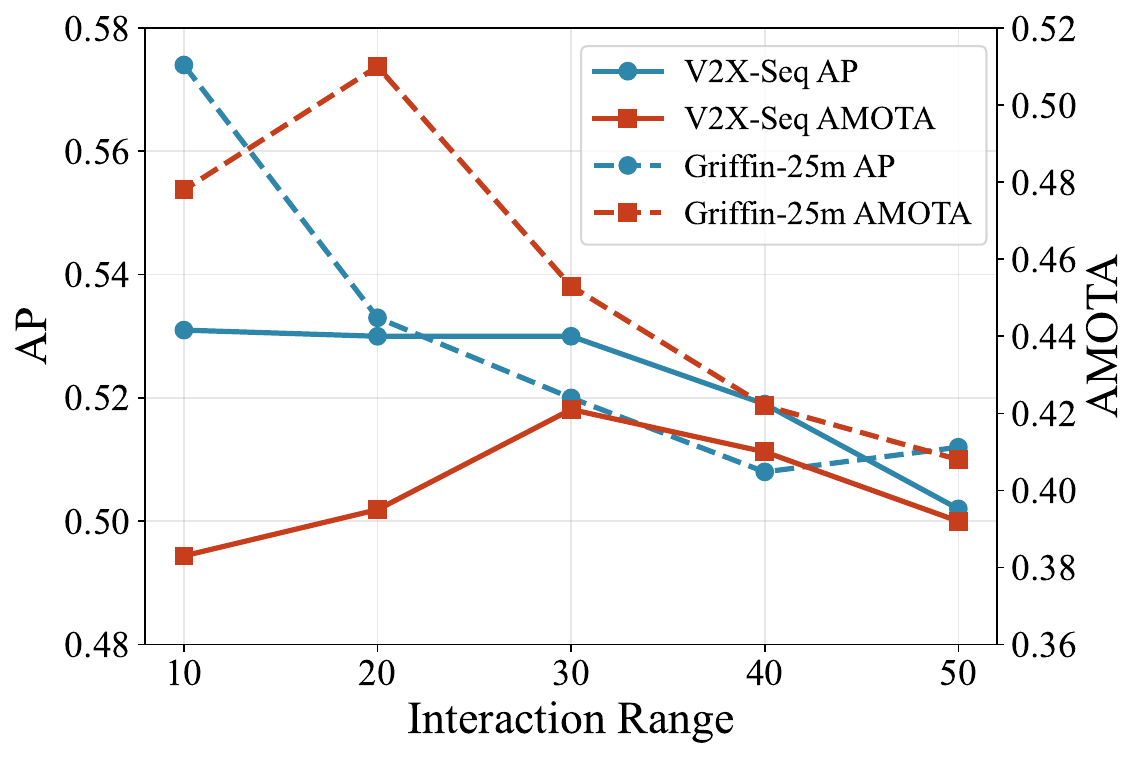}
    \caption{Effect of interaction range on two datasets.
    }
    \label{fig:interaction_range}
\end{figure}

\noindent\textbf{Effect of Interaction Range.}
A crucial hyperparameter in our framework is the interaction range ($R_{int}$), which determines the threshold for fusing cooperative instances with the ego-vehicle's perception.
As shown in Figure~\ref{fig:interaction_range}, our experiments reveal a clear and significant trend across both datasets.
As $R_{int}$ is decreased, we observe a general increase in detection accuracy (AP).
Concurrently, tracking performance (AMOTA) first improves and then declines after reaching an optimal point.
Based on our tests, the peak performance on the V2X-Seq dataset was achieved at $R_{int}=30m$, while the optimal range for the Griffin-25m dataset was found to be $15m$.

The observed increase in AP with a smaller $R_{int}$ suggests that for distant or occluded targets where the ego-vehicle's perception is inherently unreliable, forcing an interaction can be counterproductive.
The model appears to get confused by the low-quality local data, which can suppress the high-quality cooperative detection.
By directly outputting these instances, their integrity is preserved.
The behavior of AMOTA, however, indicates a trade-off.
While focusing fusion on nearby objects is beneficial, an overly restrictive $R_{int}$ can cause the system to output duplicate detections for the same object—one from the ego-vehicle and another from the cooperative agent—thereby degrading tracking performance.
Thus, the optimal $R_{int}$ represents a balance between maximizing information gain from fusion in the reliable near-field while preventing data corruption and redundancy in the far-field.

\begin{figure}[t]
    \centering
    \includegraphics[width=\columnwidth]{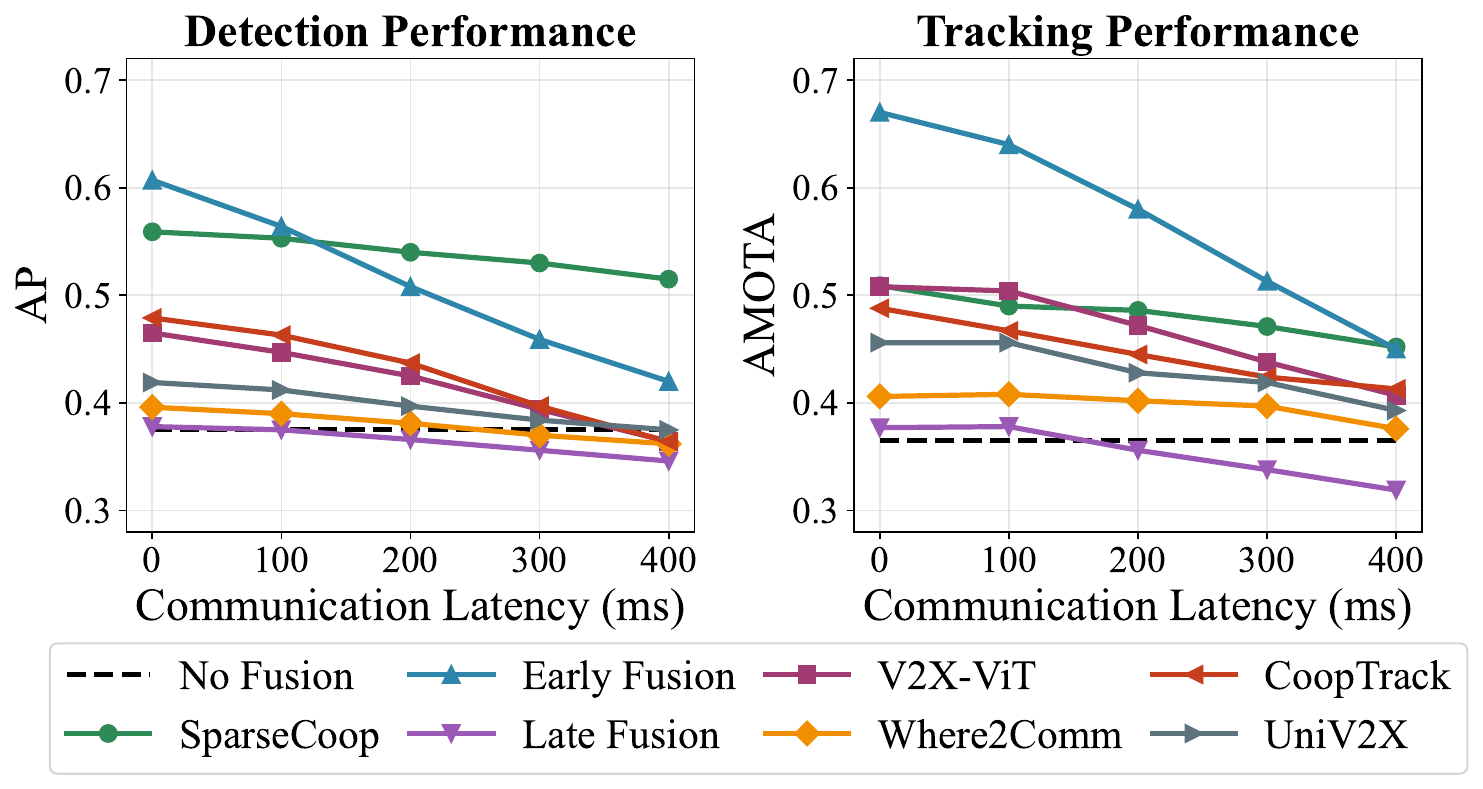}
    \caption{Impact of latency on Griffin-25m dataset.
    }
    \label{fig:communication_interference}
\end{figure}

\noindent\textbf{Robustness to Communication Latency.}
We evaluated the framework's resilience to communication latency, a critical factor for real-world deployment, using the Griffin-25m benchmark.
Results are shown in Figure~\ref{fig:communication_interference}.
While the performance of all tested cooperative methods degraded with increasing latency, SparseCoop demonstrated a markedly superior level of robustness.
At zero latency, early fusion methods held a performance advantage, as expected.
However, as latency was increased to 200ms and beyond, SparseCoop's AP scores surpassed those of all other methods, including early fusion.
This indicates a significantly more graceful performance degradation compared to competing approaches.

This exceptional robustness can be directly attributed to our framework's architectural design.
The Kinematic-Grounded Association module incorporates an explicit Latency Compensation mechanism.
By leveraging the velocity information encoded within each instance's state vector, the model applies kinematic priors to accurately predict an object's state at the ego-vehicle's current timestamp, effectively neutralizing the temporal asynchrony caused by communication delays.
In contrast, methods relying on more abstract representations like dense BEV maps lack a comparably direct and effective mechanism for temporal alignment, leading to their more pronounced performance decay.

\section{Conclusion}
\label{sec:conclusion}

In this work, we have presented SparseCoop, a fully sparse framework that solves key challenges in cooperative perception by avoiding dense BEV maps.
Our method uses kinematic-grounded queries for robust spatio-temporal alignment, a coarse-to-fine aggregation module to effectively fuse instance data, and a cooperative denoising task to stabilize training.
Experiments show SparseCoop achieves SOTA performance on the V2X-Seq and Griffin datasets.




\bibliography{THICV}

\end{document}